\documentclass[11pt,a4paper]{article}
\usepackage[hyperref]{emnlp2018}
\usepackage{times}
\usepackage{latexsym}

\usepackage{booktabs}
\usepackage{multirow}
\usepackage{graphicx}
\usepackage{array}
\usepackage{pgfplots}

\usepackage{url}

\usepackage{amsmath}
\usepackage{algorithm}
\usepackage{algorithmic}

\aclfinalcopy 

\newcommand\datasetname{\textsc{HotpotQA}}

\newcommand\squad{{SQuAD}}

\newcommand\triviaqa{{TriviaQA}}
\newcommand\searchqa{{SearchQA}}
\newcommand\wikihop{{QAngaroo}}
\newcommand\complexwebq{\textsc{ComplexWebQuestions}}
\newcommand\parlai{{ParlAI}}

\newcommand\fone{F\textsubscript{1}}

\newcommand{\bridge}[1]{{\color{orange} \textbf{\textit{#1}}}}
\newcommand{\support}[1]{{\color{blue}\textit{#1}}}
\newcommand{\answer}[1]{{\color{green!55!black}\textbf{#1}}}
\newcommand{\sentid}[1]{{\color{green!55!black}#1}}

\title{\datasetname: A Dataset for Diverse, Explainable\\
Multi-hop Question Answering}

\author{
  Zhilin Yang*$^{\spadesuit}$\quad Peng Qi*$^{\heartsuit}$ \quad Saizheng Zhang*$^{\clubsuit}$ \\
  \textbf{Yoshua Bengio$^{\clubsuit\diamondsuit}$ \quad William W. Cohen$^{\dagger}$} \quad
  \textbf{Ruslan Salakhutdinov$^{\spadesuit}$ \quad Christopher D. Manning$^{\heartsuit}$} \\
  {$\spadesuit$ Carnegie Mellon University \quad $\heartsuit$ Stanford University \quad $\clubsuit$ Mila, Universit\'e de Montr\'eal}\\
  {$\diamondsuit$ CIFAR Senior Fellow \quad $\dagger$ Google AI} \\
  {\tt \{zhiliny, rsalakhu\}@cs.cmu.edu, \{pengqi, manning\}@cs.stanford.edu}\\
  {\tt saizheng.zhang@umontreal.ca, yoshua.bengio@gmail.com, wcohen@google.com} \\
}
\date{}

\begin{document}
\maketitle

\renewcommand{\thefootnote}{\fnsymbol{footnote}}
\footnotetext[1]{These authors contributed equally. The order of authorship is decided through dice rolling.}
\footnotetext[2]{Work done when WWC was at CMU.}
\renewcommand{\thefootnote}{\arabic{footnote}}

\begin{abstract}
Existing question answering (QA) datasets fail to train QA systems to perform complex reasoning and provide explanations for answers.
We introduce \datasetname
, a new dataset with 113k Wikipedia-based question-answer pairs
with four key features:
(1) the questions require finding and reasoning over multiple supporting documents to answer;
(2) the questions are diverse and not constrained to any pre-existing knowledge bases or knowledge schemas;
(3) we provide sentence-level supporting facts required for reasoning, allowing QA systems to reason with strong supervision and explain the predictions;
(4) we offer a new type of factoid comparison questions to test QA systems' ability to extract relevant facts and perform necessary comparison.
We show that \datasetname{} is challenging for the latest QA systems, and the supporting facts enable models to improve performance and make explainable predictions.
\end{abstract}

\section{Introduction}

\begin{figure}[!t]
    \framebox{
    \parbox{0.45\textwidth}{
    \small
    \textbf{Paragraph A, Return to Olympus:} \newline
    \sentid{[1]} \support{Return to Olympus is the only album by the alternative rock band Malfunkshun.} \sentid{[2]} \support{It was released after the band had broken up and after lead singer Andrew Wood (later of Mother Love Bone) had died of a drug overdose in 1990.} \sentid{[3]} Stone Gossard, of Pearl Jam, had compiled the songs and released the album on his label, Loosegroove Records. \smallskip\newline
    \textbf{Paragraph B, Mother Love Bone:} \newline
    \sentid{[4]} \support{Mother Love Bone was an American rock band that formed in Seattle, Washington in 1987.} \sentid{[5]} The band was active from 1987 to 1990. \sentid{[6]} \support{Frontman Andrew Wood's personality and compositions helped to catapult the group to the top of the burgeoning late 1980s/early 1990s Seattle music scene.} \sentid{[7]} \support{Wood died only days before the scheduled release of the band's debut album, ``Apple'', thus ending the group's hopes of success.} \sentid{[8]} The album was finally released a few months later. \smallskip\newline
    \textbf{Q:} What was the former band of the member of Mother Love Bone who died just before the release of ``Apple''? \newline
    \textbf{A:} Malfunkshun \newline
    \textbf{Supporting facts:} \sentid{1, 2, 4, 6, 7}}
    }
    \caption{An example of the multi-hop questions in \datasetname{}. We also highlight the supporting facts in \support{blue italics}, which are also part of the dataset.} \label{fig:example}
\end{figure}

The ability to perform reasoning and inference over natural language is an important aspect of intelligence.
The task of question answering (QA) provides a quantifiable and objective way to test the reasoning ability of intelligent systems.
To this end, a few large-scale QA datasets have been proposed, which sparked significant progress in this direction.
However, existing datasets have limitations that hinder further advancements of machine reasoning over natural language, especially in testing QA systems' ability to perform \emph{multi-hop reasoning}, where the system has to reason with information taken from more than one document to arrive at the answer.

First, some datasets mainly focus on testing the ability of reasoning within a single paragraph or document, or \emph{single-hop reasoning}.
For example, in \squad{} \cite{rajpurkar2016squad} questions are designed to be answered given a single paragraph as the context, and most of the questions can in fact be answered by matching the question with a single sentence in that paragraph.
As a result, it has fallen short at testing systems' ability to reason over a larger context.
\triviaqa{} \cite{Joshi2017TriviaQA} and \searchqa{} \cite{dunn2017searchqa} create a more challenging setting by using information retrieval to collect multiple documents to form the context given existing question-answer pairs.
Nevertheless, most of the questions can be answered by matching the question with a few nearby sentences in one single paragraph, which is limited as it does not require more complex reasoning (e.g., over multiple paragraphs).

Second, existing datasets that target multi-hop reasoning, such as \wikihop{} \cite{welbl2017constructing} and \complexwebq{} \cite{talmor2018web}, are constructed using existing knowledge bases (KBs). As a result, these datasets are constrained by the schema of the KBs they use, and therefore the diversity of questions and answers is inherently limited.

Third, all of the above datasets only provide distant supervision; i.e., the systems only know what the answer is, but do not know what supporting facts lead to it. This makes it difficult for models to learn about the underlying reasoning process, as well as to make explainable predictions.

To address the above challenges, we aim at creating a QA dataset that requires reasoning over multiple documents, and does so in natural language, without constraining itself to an existing knowledge base or knowledge schema.
We also want it to provide the system with strong supervision about what text the answer is actually derived from, to help guide systems to perform meaningful and explainable reasoning.

We present \datasetname%
\footnote{The name comes from the first three authors' arriving at the main idea during a discussion at a hot pot restaurant.}%
, a large-scale dataset that satisfies these desiderata.
\datasetname{} is collected by crowdsourcing based on Wikipedia articles, where crowd workers are shown multiple supporting context documents and asked explicitly to come up with questions requiring reasoning about all of the documents.
This ensures it covers multi-hop questions that are more natural, and are not designed with any pre-existing knowledge base schema in mind.
Moreover, we also ask the crowd workers to provide the supporting facts they use to answer the question, which we also provide as part of the dataset (see Figure \ref{fig:example} for an example).
We have carefully designed a data collection pipeline for \datasetname, since  the collection of high-quality multi-hop questions is non-trivial.
We hope that this pipeline also sheds light on future work in this direction.
Finally, we also collected a novel type of questions---comparison questions---as part of \datasetname, in which we require systems to compare two entities on some shared properties to test their understanding of both language and common concepts such as numerical magnitude. We make \datasetname{} publicly available at \href{https://hotpotqa.github.io}{\bf https://HotpotQA.github.io}.

\section{Data Collection} \label{sec:collect}


The main goal of our work is to collect a diverse and explainable question answering dataset that requires multi-hop reasoning.
One way to do so is to define reasoning chains based on a knowledge base \cite{welbl2017constructing,talmor2018web}.
However, the resulting datasets are limited by the incompleteness of entity relations and the lack of diversity in the question types.
Instead, in this work, we focus on text-based question answering in order to diversify the questions and answers.
The overall setting is that given some context paragraphs (e.g., a few paragraphs, or the entire Web) and a question, a QA system answers the question by extracting a span of text from the context, similar to \citet{rajpurkar2016squad}.
We additionally ensure that it is necessary to perform multi-hop reasoning to correctly answer the question.

It is non-trivial to collect text-based multi-hop questions.
In our pilot studies, we found that simply giving an arbitrary set of paragraphs to crowd workers is counterproductive, because for most paragraph sets, it is difficult to ask a meaningful multi-hop question.
To address this challenge, we carefully design a pipeline to collect text-based multi-hop questions.
Below, we will highlight the key design choices in our pipeline.

\paragraph{Building a Wikipedia Hyperlink Graph.}
We use the entire English Wikipedia dump as our corpus.\footnote{\url{https://dumps.wikimedia.org/}}
In this corpus, we make two observations:
(1) hyper-links in the Wikipedia articles often naturally entail a relation between two (already disambiguated) entities in the context, which could potentially be used to facilitate multi-hop reasoning;
(2) the first paragraph of each article often contains much information that could be queried in a meaningful way.
Based on these observations, we extract all the hyperlinks from the first paragraphs of all Wikipedia articles.
With these hyperlinks, we build a directed graph $G$, where each edge $(a, b)$ indicates there is a hyperlink from the first paragraph of article $a$ to article $b$.

\paragraph{Generating Candidate Paragraph Pairs.}
To generate meaningful pairs of paragraphs for multi-hop question answering with $G$, we start by considering an example question ``when was the singer and songwriter of Radiohead born?''
To answer this question, one would need to first reason that the ``singer and songwriter of Radiohead'' is ``Thom Yorke'', and then figure out his birthday in the text.
We call ``Thom Yorke'' a \emph{bridge entity} in this example.
Given an edge $(a, b)$ in the hyperlink graph $G$, the entity of $b$ can usually be viewed as a bridge entity that connects $a$ and $b$.
As we observe articles $b$ usually determine the theme of the shared context between $a$ and $b$, but not all articles $b$ are suitable for collecting multi-hop questions.
For example, entities like countries are frequently referred to in Wikipedia, but don't necessarily have much in common with all incoming links.
It is also difficult, for instance, for the crowd workers to ask meaningful multi-hop questions about highly technical entities like the IPv4 protocol.
To alleviate this issue, we constrain the bridge entities to a set of manually curated pages in Wikipedia (see Appendix \ref{sec:collection_details}).
After curating a set of pages $B$, we create candidate paragraph pairs by sampling edges $(a, b)$ from the hyperlink graph such that $b \in B$.

\paragraph{Comparison Questions.}
In addition to questions collected using bridge entities, we also collect another type of multi-hop questions---comparison questions.
The main idea is that comparing two entities from the same category usually results in interesting multi-hop questions, e.g., ``Who has played for more NBA teams, Michael Jordan or Kobe Bryant?''
To facilitate collecting this type of question, we manually curate 42 lists of similar entities (denoted as $L$) from Wikipedia.\footnote{This is achieved by manually curating lists from the Wikipedia ``List of lists of lists'' (\url{https://wiki.sh/y8qv}). One example is ``Highest Mountains on Earth''.}
To generate candidate paragraph pairs, we randomly sample two paragraphs from the same list and present them to the crowd worker.

To increase the diversity of multi-hop questions, we also introduce a subset of yes/no questions in comparison questions.
This complements the original scope of comparison questions by offering new ways to require systems to reason over both paragraphs.
For example, consider the entities Iron Maiden (from the UK) and AC/DC (from Australia).
Questions like ``Is Iron Maiden or AC/DC from the UK?'' are not ideal, because one would deduce the answer is ``Iron Maiden'' even if one only had access to that article.
With yes/no questions, one may ask ``Are Iron Maiden and AC/DC from the same country?'', which requires reasoning over both paragraphs.

To the best of our knowledge, text-based comparison questions are a novel type of questions that have not been considered by previous datasets.
More importantly, answering these questions usually requires arithmetic comparison, such as comparing ages given birth dates, which presents a new challenge for future model development.

\paragraph{Collecting Supporting Facts.}
To enhance the explainability of question answering systems, we want them to output a set of \emph{supporting facts} necessary to arrive at the answer, when the answer is generated.
To this end, we also collect the sentences that determine the answers from crowd workers.
These supporting facts can serve as strong supervision for what sentences to pay attention to.
Moreover, we can now test the explainability of a model by comparing the predicted supporting facts to the ground truth ones.

The overall procedure of data collection is illustrated in Algorithm \ref{algo:collect}.

\begin{algorithm}[t]
\small
    \caption{Overall data collection procedure}
    \label{algo:collect}
\begin{algorithmic}
    \STATE {\bfseries Input:} question type ratio $r_1=0.75$, yes/no ratio $r_2=0.5$
    \WHILE{not finished}
        \IF {$\mathrm{random()} < r_1$}
            \STATE Uniformly sample an entity $b \in B$
            \STATE Uniformly sample an edge $(a, b)$
            \STATE Workers ask a question about paragraphs $a$ and $b$
        \ELSE
            \STATE Sample a list from $L$, with probabilities weighted by list sizes
            \STATE Uniformly sample two entities $(a, b)$ from the list
            \IF {$\mathrm{random()} < r_2$}
                \STATE Workers ask a yes/no question to compare $a$ and $b$
            \ELSE
                \STATE Workers ask a question with a span answer to compare $a$ and $b$
            \ENDIF
        \ENDIF
        \STATE Workers provide the supporting facts
    \ENDWHILE
\end{algorithmic}
\end{algorithm}






\section{Processing and Benchmark Settings}

We collected 112,779 valid examples in total on Amazon Mechanical Turk\footnote{\url{https://www.mturk.com/}} using the ParlAI interface \cite{miller2017parlai} (see Appendix \ref{sec:collection_details}).%
To isolate potential single-hop questions from the desired multi-hop ones, we first split out a subset of data called \textit{train-easy}.
Specifically, we randomly sampled questions ($\sim$3--10 per Turker) from top-contributing turkers, and categorized all their questions into the \textit{train-easy} set if an overwhelming percentage in the sample only required reasoning over one of the paragraphs.
We sampled these turkers because they contributed more than 70\% of our data.
This \textit{train-easy} set contains 18,089 mostly single-hop examples.

We implemented a question answering model based on the current state-of-the-art architectures, which we discuss in detail in Section \ref{sec:baseline}.
Based on this model, we performed a three-fold cross validation on the remaining multi-hop examples.
Among these examples, the models were able to correctly answer 60\% of the questions with high confidence (determined by thresholding the model loss).
These correctly-answered questions (56,814 in total, 60\% of the multi-hop examples) are split out and marked as the \textit{train-medium} subset, which will also be used as part of our training set.

After splitting out \textit{train-easy} and \textit{train-medium}, we are left with hard examples.
As our ultimate goal is to solve multi-hop question answering, we focus on questions that the latest modeling techniques are not able to answer.
Thus we constrain our dev and test sets to be hard examples.
Specifically, we randomly divide the hard examples into four subsets, \textit{train-hard}, \textit{dev}, \textit{test-distractor}, and \textit{test-fullwiki}.
Statistics about the data split can be found in Table \ref{tab:split}. In Section \ref{sec:exp}, we will show that combining \textit{train-easy}, \textit{train-medium}, and \textit{train-hard} to train models yields the best performance, so we use the combined set as our default training set. The two test sets \textit{test-distractor} and \textit{test-fullwiki} are used in two different benchmark settings, which we introduce next.

\begin{table}
\small
    \centering
    \begin{tabular}{l l l r}
        \toprule
        Name & Desc. & Usage & \# Examples \\
        \midrule
        train-easy & single-hop & training & 18,089 \\
        train-medium & multi-hop & training & 56,814 \\
        train-hard & hard multi-hop & training & 15,661 \\
        dev & hard multi-hop & dev & 7,405 \\
        test-distractor & hard multi-hop & test & 7,405 \\
        test-fullwiki & hard multi-hop & test & 7,405 \\
        \multicolumn{3}{l}{Total} & 112,779 \\
        \bottomrule
    \end{tabular}
    \caption{Data split. The splits \textit{train-easy}, \textit{train-medium}, and \textit{train-hard} are combined for training. The distractor and full wiki settings use different test sets 
    so that the gold paragraphs in the full wiki test set remain unknown to any models.} \label{tab:split}
\end{table}

We create two benchmark settings.
In the first setting, to challenge the model to find the true supporting facts in the presence of noise, for each example we employ bigram tf-idf \cite{chen2017reading} to retrieve 8 paragraphs from Wikipedia as \textit{distractors}, using the question as the query.
We mix them with the 2 gold paragraphs (the ones used to collect the question and answer) to construct the \textbf{distractor} setting.
The 2 gold paragraphs and the 8 distractors are shuffled before they are fed to the model.
In the second setting, we fully test the model's ability to locate relevant facts as well as reasoning about them by requiring it to answer the question given the first paragraphs of all Wikipedia articles without the gold paragraphs specified.
This \textbf{full wiki} setting truly tests the performance of the systems' ability at multi-hop reasoning in the wild.\footnote{As we required the crowd workers to use complete entity names in the question, the majority of the questions are unambiguous in the full wiki setting.}
The two settings present different levels of difficulty, and would require techniques ranging from reading comprehension to information retrieval.
As shown in Table \ref{tab:split}, we use separate test sets for the two settings to avoid leaking information, because the gold paragraphs are available to a model in the distractor setting, but should not be accessible in the full wiki setting.

We also try to understand the model's good performance on the \textit{train-medium} split.
Manual analysis shows that the ratio of multi-hop questions in \textit{train-medium} is similar to that of the hard examples (93.3\% in \textit{train-medium} vs.\ 92.0\% in \textit{dev}), but one of the question types appears more frequently in \textit{train-medium} compared to the hard splits (Type II: 32.0\% in \textit{train-medium} vs.\ 15.0\% in \textit{dev}, see Section \ref{sec:analysis} for the definition of Type II questions).
These observations demonstrate that given enough training data, existing neural architectures can be trained to answer certain types and certain subsets of the multi-hop questions.
However, \textit{train-medium} remains challenging when not just the gold paragraphs are present---we show in Appendix \ref{sec:fullwiki_details} that the retrieval problem on these examples are as difficult as that on their hard cousins.


\section{Dataset Analysis} \label{sec:analysis}

In this section, we analyze the types of questions, types of answers, and types of multi-hop reasoning covered in the dataset.

\begin{figure}[!t]
    \centering
    \includegraphics[width=.48\textwidth]{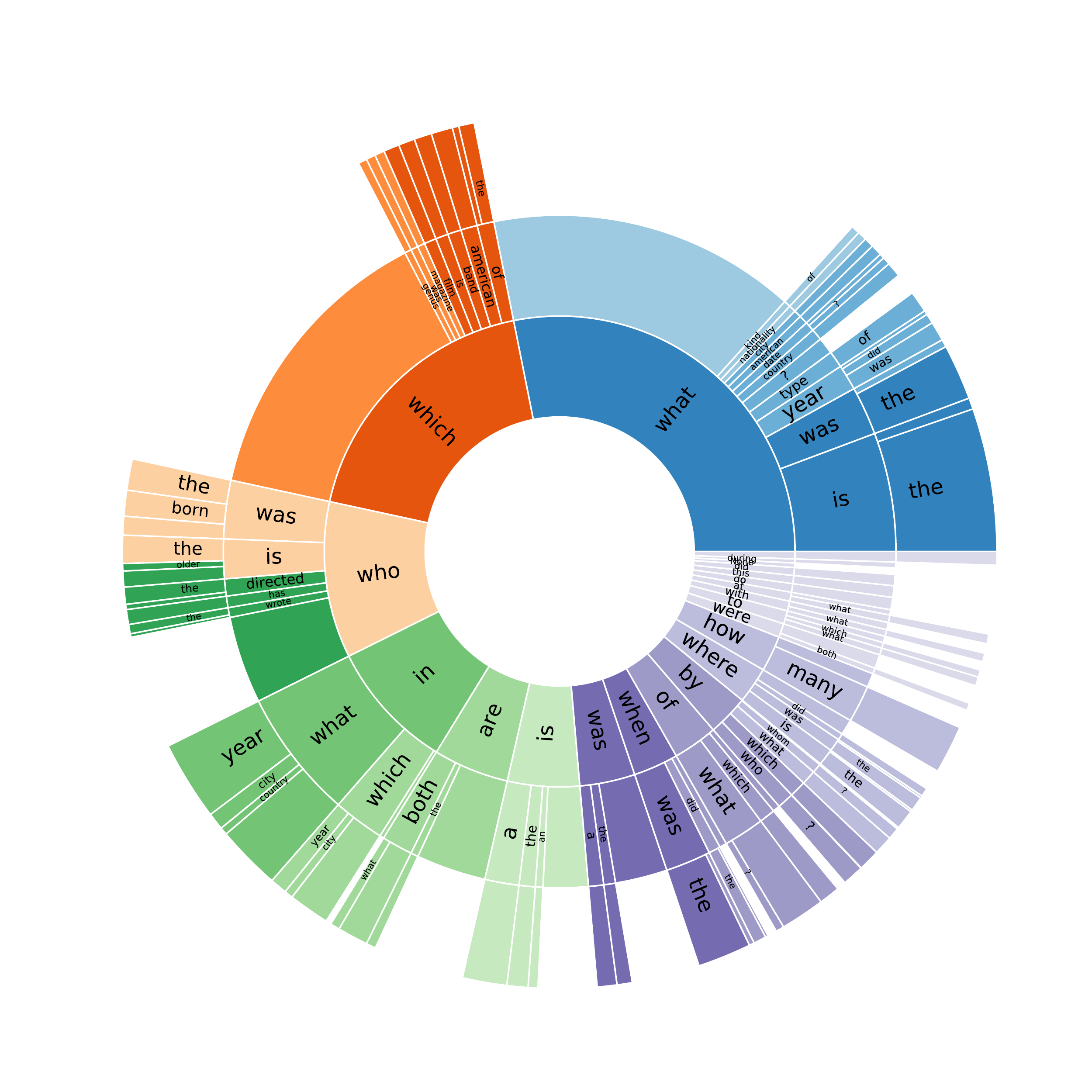}
    \vskip -0.7em
    \caption{Types of questions covered in \datasetname{}. Question types are extracted heuristically, starting at question words or prepositions  preceding them. Empty colored blocks indicate suffixes that are too rare to show individually. See main text for more details.} \label{fig:question_types}
\end{figure}

\paragraph{Question Types.}
We heuristically identified question types for each collected question.
To identify the question type, we first locate the \emph{central question word} (CQW) in the question.
Since \datasetname{} contains comparison questions and yes/no questions, we consider as \emph{question words} WH-words, copulas (``is'', ``are''), and auxiliary verbs (``does'', ``did'').
Because questions often involve relative clauses beginning with WH-words, we define the CQW as the first question word in the question if it can be found in the first three tokens, or the last question word otherwise.
Then, we determine question type by extracting words up to 2 tokens away to the right of the CQW, along with the token to the left if it is one of a few common prepositions (e.g., in the cases of  ``in which'' and ``by whom'').

We visualize the distribution of question types in Figure \ref{fig:question_types}, and label the ones shared among more than 250 questions.
As is shown, our dataset covers a diverse variety of questions centered around entities, locations, events, dates, and numbers, as well as yes/no questions directed at comparing two entities (``Are both A and B ...?''), to name a few.

\paragraph{Answer Types.}
We further sample 100 examples from the dataset, and present the types of answers in Table \ref{tab:answer_types}.
As can be seen, \datasetname{} covers a broad range of answer types, which matches our initial analysis of question types.
We find that a majority of the questions are about entities in the articles (68\%), and a non-negligible amount of questions also ask about various properties like date (9\%) and other descriptive properties such as numbers (8\%) and adjectives (4\%).

\begin{table}[!t]
    \centering
    \small
    \begin{tabular}{p{2cm}rp{4.2cm}}
        \toprule
        Answer Type & \% & Example(s) \\
        \midrule
        Person & 30 & King Edward II, Rihanna \\
        Group / Org & 13 & Cartoonito, Apalachee \\
        Location & 10 & Fort Richardson, California \\
        Date & 9 & 10th or even 13th century\\
        Number & 8 & 79.92 million, 17 \\
        Artwork & 8 & Die schweigsame Frau \\
        Yes/No & 6 & - \\
        Adjective & 4 & conservative \\
        Event & 1 & Prix Benois de la Danse \\
        Other proper noun & 6 & Cold War, Laban Movement Analysis \\
        Common noun & 5 & comedy, both men and women\\
        \bottomrule
    \end{tabular}
    \vskip -0.7em
    \caption{Types of answers in \datasetname{}.} \label{tab:answer_types}
\end{table}

\begin{table*}[!t]
    \centering
    \small
    \begin{tabular}{p{2.9cm}rp{11.7cm}}
        \toprule
        Reasoning Type & \% & Example(s) \\
        \midrule
        Inferring the \bridge{bridge entity} to complete the 2nd-hop question (Type I) & 42 &
        \textbf{Paragraph A:} The 2015 Diamond Head Classic was a college basketball tournament ...  \bridge{Buddy Hield} \support{was named the tournament's MVP}.\newline
        \textbf{Paragraph B:} \bridge{Chavano Rainier "Buddy" Hield} is a Bahamian professional basketball player for the \answer{Sacramento Kings} of the NBA... \newline
        \textbf{Q:}  Which team does the player named 2015 Diamond Head Classic's MVP play for?
        \\
        \midrule
        Comparing two entities (Comparison) & 27 &
        \textbf{Paragraph A:} LostAlone were a British rock band ... consisted of \support{Steven Battelle, Alan Williamson, and Mark Gibson}... \newline
        \textbf{Paragraph B:} Guster is an American alternative rock band ... Founding members \support{Adam Gardner, Ryan Miller, and Brian Rosenworcel} began...\newline
        \textbf{Q:}  Did LostAlone and Guster have the same number of members? (\answer{yes})
        \\
        \midrule
        Locating the \answer{answer entity} by checking multiple properties (Type II) & 15 &
        \textbf{Paragraph A:} Several \support{current and former members of the Pittsburgh Pirates} – ... John Milner, \answer{Dave Parker}, and Rod Scurry...\newline
        \textbf{Paragraph B:} \answer{David Gene Parker}, \support{nicknamed "The Cobra"}, is an American former player in Major League Baseball...\newline
        \textbf{Q:} Which former member of the Pittsburgh Pirates was nicknamed "The Cobra"?
         \\
         \midrule
         Inferring about the property of an entity in question through a \bridge{bridge entity} (Type III) & 6 &
         \textbf{Paragraph A:}
         \support{Marine Tactical Air Command Squadron 28} is a United States Marine Corps aviation command and control unit based at \bridge{Marine Corps Air Station Cherry Point}...\newline
         \textbf{Paragraph B:} \bridge{Marine Corps Air Station Cherry Point} ... is a United States Marine Corps airfield located in \answer{Havelock, North Carolina}, USA ...\newline
         \textbf{Q:} What city is the Marine Air Control Group 28 located in?\\
         \midrule
         Other types of reasoning that require more than two supporting facts (Other) & 2 &
        \textbf{Paragraph A:} ... the towns of Yodobashi, \answer{Okubo, Totsuka, and Ochiai town} \support{were merged into Yodobashi ward}. ... \bridge{Yodobashi Camera} \support{is a store with its name taken from the town and ward}.\newline
        \textbf{Paragraph B}: \bridge{Yodobashi Camera} Co., Ltd. is \support{a major Japanese retail chain specializing in electronics, PCs, cameras and photographic equipment}.\newline
        \textbf{Q:} Aside from Yodobashi, what other towns were merged into the ward which gave the major Japanese retail chain specializing in electronics, PCs, cameras, and photographic equipment it's name?
         \\
        \bottomrule
    \end{tabular}
    \vskip -0.7em
    \caption{Types of multi-hop reasoning required to answer questions in the \datasetname{} dev and test sets.
    We show in \bridge{orange bold italics} bridge entities if applicable, \support{blue italics} supporting facts from the paragraphs that connect directly to the question, and \answer{green bold} the answer in the paragraph or following the question.
    The remaining 8\% are single-hop (6\%) or unanswerable questions (2\%) by our judgement.
    } \label{tab:multihop_types}
\end{table*}

\paragraph{Multi-hop Reasoning Types.}
We also sampled 100 examples from the dev and test sets and manually classified the types of reasoning required to answer each question.
Besides comparing two entities, there are three main types of multi-hop reasoning required to answer these questions, which we show in Table \ref{tab:multihop_types} accompanied with examples.

Most of the questions  require at least one supporting fact from each paragraph to answer.
A majority of sampled questions (42\%) require chain reasoning (Type I in the table), where the reader must first identify a bridge entity before the second hop can be answered by filling in the bridge.
One strategy to answer these questions would be to decompose them into consecutive single-hop questions.
The bridge entity could also be used implicitly to help infer properties of other entities related to it.
In some questions (Type III), the entity in question shares certain properties with a bridge entity (e.g., they are collocated), and we can infer its properties through the bridge entity.
Another type of question involves locating the answer entity by satisfying multiple properties simultaneously (Type II).
Here, to answer the question, one could find the set of all entities that satisfy each of the properties mentioned, and take an intersection to arrive at the final answer.
Questions comparing two entities (Comparison) also require the system to understand the properties in question about the two entities (e.g., nationality), and sometimes require arithmetic such as counting (as seen in the table) or comparing numerical values (``Who is older, A or B?'').
Finally, we find that sometimes the questions require more than two supporting facts to answer (Other).
In our analysis, we also find that for all of the examples shown in the table, the supporting facts provided by the
Turkers match exactly with the limited context shown here, showing that the supporting facts collected are of high quality.

Aside from the reasoning types mentioned above, we also estimate that about 6\% of the sampled questions can be answered with one of the two paragraphs, and 2\% of them unanswerable.
We also randomly sampled 100 examples from \emph{train-medium} and \emph{train-hard} combined, and the proportions of reasoning types are: Type I 38\%, Type II 29\%, Comparison 20\%, Other 7\%, Type III 2\%, single-hop 2\%, and unanswerable 2\%.


\section{Experiments} \label{sec:exp}

\begin{figure}[!h]
    \centering
    \includegraphics[width=.45\textwidth]{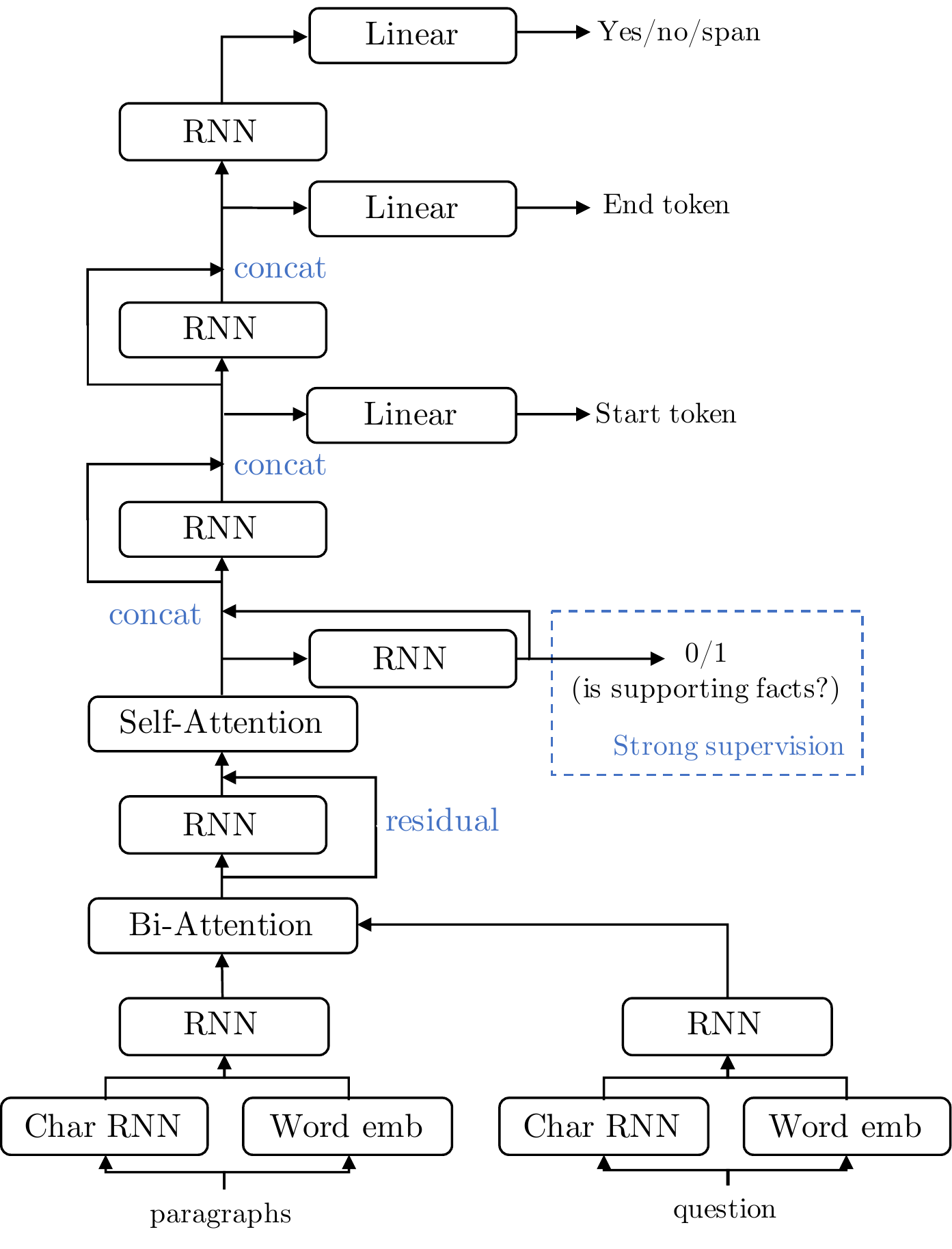}
    \vskip -1em
    \caption{Our model architecture. Strong supervision over supporting facts is used in a multi-task setting.} \label{fig:model}
\end{figure}

\subsection{Model Architecture and Training} \label{sec:baseline}

To test the performance of leading QA systems on our data, we reimplemented the architecture described in \citet{clark2017simple} as our baseline model. We note that our implementation without weight averaging achieves performance very close to what the authors reported on \squad{} (about 1 point worse in \fone{}). Our implemented model subsumes the latest technical advances on question answering, including character-level models, self-attention \cite{wang2017gated}, and bi-attention \cite{seo2016bidirectional}.
Combining these three key components is becoming standard practice, and various state-of-the-art or competitive architectures \cite{liu2017stochastic,clark2017simple,wang2017gated,seo2016bidirectional,pan2017memen,salant2017contextualized,xiong2017dcn+} on SQuAD can be viewed as similar to our implemented model.
To accommodate yes/no questions, we also add a 3-way classifier after the last recurrent layer to produce the probabilities of ``yes'', ``no'', and span-based answers.
During decoding, we first use the 3-way output to determine whether the answer is ``yes'', ``no'', or a text span. If it is a text span, we further search for the most probable span.



\paragraph{Supporting Facts as Strong Supervision.}
To evaluate the baseline model's performance in predicting explainable supporting facts, as well as how much they improve QA performance, we additionally design a component to incorporate such strong supervision into our model.
For each sentence, we concatenate the output of the self-attention layer at the first and last positions, and use a binary linear classifier to predict the probability that the current sentence is a supporting fact.
We minimize a binary cross entropy loss for this classifier.
This 
objective is jointly optimized with the normal question answering objective in a multi-task learning setting, and they share the same low-level representations.
With this classifier, the model can also be evaluated on the task of supporting fact prediction to gauge its explainability.
Our overall architecture is illustrated in Figure \ref{fig:model}.
Though it is possible to build a pipeline system, in this work we focus on an end-to-end one, which is easier to tune and faster to train.



\subsection{Results}

\begin{table*}
    \small
    \centering
    \begin{tabular}{llcccccc}
        \toprule
        \multirow{2}{*}{Setting} & \multirow{2}{*}{Split} & \multicolumn{2}{c}{Answer} & \multicolumn{2}{c}{Sup Fact} & \multicolumn{2}{c}{Joint} \\
        \cmidrule{3-8}
        & & EM & \fone & EM & \fone & EM & \fone \\
        \midrule
        distractor & dev & 44.44 & 58.28 & 21.95 & 66.66 & 11.56 & 40.86 \\
        distractor & test & 45.46 & 58.99 & 22.24 & 66.62 & 12.04 & 41.37 \\
        \midrule
        full wiki & dev & 24.68 & 34.36 & \phantom{0}5.28 & 40.98 & \phantom{0}2.54 & 17.73 \\
        full wiki & test & 25.23 & 34.40 & \phantom{0}5.07 & 40.69 & \phantom{0}2.63 & 17.85 \\
        \bottomrule
    \end{tabular}
    \caption{Main results: the performance of question answering and supporting fact prediction in the two benchmark settings. We encourage researchers to report these metrics when evaluating their methods.} \label{tab:main}
    \vskip -0.7em
\end{table*}

\begin{table}[t]
    \small
    \centering
    \begin{tabular}{lcccc}
        \toprule
        Set & MAP & Mean Rank & Hits@2 & Hits@10 \\
        \midrule
        dev & 43.93 & 314.71& 39.43 & 56.06  \\
        test & 43.21 & 314.05& 38.67 & 55.88 \\
        \bottomrule
    \end{tabular}
    \caption{Retrieval performance in the full wiki setting. Mean Rank is averaged over the ranks of two gold paragraphs.} \label{tab:fullwiki_ir}
    \vskip -0.2em
\end{table}



\begin{table}[t]
    \small
    \centering
    \begin{tabular}{lcccc}
        \toprule
        Setting & Br EM & Br \fone & Cp EM & Cp \fone \\
        \midrule
        distractor & 43.41 & 59.09 & 48.55 & 55.05 \\
        full wiki & 19.76 & 30.42 & 43.87 & 50.70 \\
        \bottomrule
    \end{tabular}
    \caption{Performance breakdown over different question types on the dev set in the distractor setting. ``Br'' denotes questions collected using bridge entities, and ``Cp'' denotes comparison questions.} \label{tab:break}
    \vskip -0.2em
\end{table}


\begin{table}[t]
    \small
    \centering
    \begin{tabular}{lcc}
        \toprule
        Setting & EM & \fone \\
        \midrule
        our model & 44.44 & 58.28 \\
        \midrule
        -- sup fact & 42.79 & 56.19 \\
        \midrule
        -- sup fact, self attention & 41.59 & 55.19 \\
        -- sup fact, char model & 41.66 & 55.25 \\
        \midrule
        -- sup fact, train-easy & 41.61 & 55.12 \\
        -- sup fact, train-easy, train-medium & 31.07 & 43.61 \\
        \midrule
        gold only & 48.38 & 63.58 \\
        sup fact only & 51.95 & 66.98 \\
        \bottomrule
    \end{tabular}
    \caption{Ablation study of question answering performance on the dev set in the distractor setting. ``-- sup fact'' means removing strong supervision over supporting facts from our model. ``-- train-easy'' and ``-- train-medium'' means discarding the according data splits from training. ``gold only'' and ``sup fact only'' refer to using the gold paragraphs or the supporting facts as the only context input to the model.} \label{tab:ablation}
    \vskip -0.2em
\end{table}


We evaluate our model in the two benchmark settings.
In the full wiki setting, to enable efficient tf-idf retrieval among 5,000,000+ wiki paragraphs, given a question we first return a candidate pool of at most 5,000 paragraphs using an inverted-index-based filtering strategy\footnote{See Appendix \ref{sec:fullwiki_details} for details.} and then select the top 10 paragraphs in the pool as the final candidates using bigram tf-idf.\footnote{We choose the number of final candidates as 10 to stay consistent with the distractor setting where candidates are 2 gold paragraphs plus 8 distractors.}
Retrieval performance is shown in Table \ref{tab:fullwiki_ir}.
After retrieving these 10 paragraphs, we then use the model trained in the distractor setting to evaluate its performance on these final candidate paragraphs.


Following previous work \cite{rajpurkar2016squad}, we use exact match (EM) and \fone{} as two evaluation metrics.
To assess the explainability of the models, we further introduce two sets of metrics involving the supporting facts.
The first set focuses on evaluating the supporting facts directly, namely EM and \fone{} on the set of supporting fact sentences as compared to the gold set.
The second set features joint metrics that combine the evaluation of answer spans and supporting facts as follows.
For each example, given its precision and recall on the answer span ($P^{\text{(ans)}}, R^{\text{(ans)}}$) and the supporting facts ($P^{\text{(sup)}}, R^{\text{(sup)}}$), respectively, we calculate joint \fone{} as
\begin{align*}
P^{\text{(joint)}} = P^{\text{(ans)}}P^{\text{(sup)}}, ~~R^{\text{(joint)}} = R^{\text{(ans)}}R^{\text{(sup)}},
\end{align*}
\vspace{-2em}
\begin{align*}
\text{Joint \fone} = \frac{2 P^{\text{(joint)}} R^{\text{(joint)}}}{P^{\text{(joint)}} + R^{\text{(joint)}}}.
\end{align*}
Joint EM is 1 only if both tasks achieve an exact match and otherwise 0.
Intuitively, these metrics penalize systems that perform poorly on either task.
All metrics are evaluated example-by-example, and then averaged over examples in the evaluation set.

The performance of our model on the benchmark settings is reported in Table \ref{tab:main}, where all numbers are obtained with strong supervision over supporting facts. From the distractor setting to the full wiki setting, expanding the scope of the context increases the difficulty of question answering.
The performance in the full wiki setting is substantially lower, which poses a challenge to existing techniques on retrieval-based question answering.
Overall, model performance in all settings is significantly lower than human performance as shown in Section \ref{sec:human}, which indicates that more technical advancements are needed in future work.

We also investigate the explainability of our model by measuring supporting fact prediction performance.
Our model achieves 60+ supporting fact prediction \fone{} and $\sim$40 joint \fone{}, which indicates there is room for further improvement in terms of explainability.


In Table \ref{tab:break}, we break down the performance on different question types.
In the distractor setting, comparison questions have lower \fone{} scores than questions involving bridge entities (as defined in Section 2), which indicates that better modeling this novel question type might need better neural architectures.
In the full wiki setting, the performance of bridge entity questions drops significantly while that of comparison questions decreases only marginally.
This is because both entities usually appear in the comparison questions, and thus reduces the difficulty of retrieval.
Combined with the retrieval performance in Table \ref{tab:fullwiki_ir}, we believe that the deterioration in the full wiki setting in Table \ref{tab:main} is largely due to the difficulty of retrieving both entities.

We perform an ablation study in the distractor setting, and report the results in Table \ref{tab:ablation}.
Both self-attention and character-level models contribute notably to the final performance, which is consistent with prior work.
This means that techniques targeted at single-hop QA are still somewhat effective in our setting.
Moreover, removing strong supervision over supporting facts decreases performance, which demonstrates the effectiveness of our approach and the usefulness of the supporting facts.
We establish an estimate of the upper bound of strong supervision by only considering the supporting facts as the oracle context input to our model, which achieves a 10+ \fone{} improvement over not using the supporting facts.
Compared with the gain of strong supervision in our model ($\sim$2 points in \fone{}), our proposed method of incorporating supporting facts supervision is most likely suboptimal, and we leave the challenge of better modeling to future work.
At last, we show that combining all data splits (\textit{train-easy}, \textit{train-medium}, and \textit{train-hard}) yields the best performance, which is adopted as the default setting.


\subsection{Establishing Human Performance} \label{sec:human}

To establish human performance on our dataset, we randomly sampled 1,000 examples from the dev and test sets, and had at least three additional Turkers provide answers and supporting facts for these examples.
As a baseline, we treat the original Turker during data collection as the prediction, and the newly collected answers and supporting facts as references, to evaluate human performance.
For each example, we choose the answer and supporting fact reference that maximize the \fone{} score to report the final metrics to reduce the effect of ambiguity \cite{rajpurkar2016squad}.

As can be seen in Table \ref{tab:human_perf}, the original crowd worker achieves very high performance in both finding supporting facts, and answering the question correctly.
If the baseline model were provided with the correct supporting paragraphs to begin with, it achieves parity with the crowd worker in finding supporting facts, but still falls short at finding the actual answer.
When distractor paragraphs are present,  the performance gap between the baseline model and the crowd worker on both tasks is enlarged to $\sim$30\% for both EM and \fone{}.

We further establish the upper bound of human performance in \datasetname, by taking the maximum EM and \fone{} for each example.
Here, we use each Turker's answer in turn as the prediction, and evaluate it against all other workers' answers.
As can be seen in Table \ref{tab:human_perf}, most of the metrics are close to 100\%, illustrating that on most examples, at least a subset of Turkers agree with each other, showing high inter-annotator agreement.
We also note that crowd workers agree less on supporting facts, which could reflect that this task is inherently more subjective than answering the question.

\setlength{\tabcolsep}{4pt}
\begin{table}
    \small
    \centering
    \begin{tabular}{lcccccc}
        \toprule
        \multirow{2}{*}{Setting} & \multicolumn{2}{c}{Answer} & \multicolumn{2}{c}{Sp Fact} & \multicolumn{2}{c}{Joint} \\
        \cmidrule{2-7}
        & EM & \fone & EM & \fone & EM & \fone \\
        \midrule
        gold only & 65.87 & 74.67 & 59.76 & 90.41 & 41.54 & 68.15 \\
        distractor & 60.88 & 68.99 & 30.99 & 74.67 & 20.06 & 52.37 \\
        \midrule
        Human & 83.60 & 91.40 & 61.50 & 90.04 & 52.30 & 82.55 \\
        Human UB & 96.80 & 98.77 & 87.40 & 97.56 & 84.60 & 96.37 \\
        \bottomrule
    \end{tabular}
    \caption{Comparing baseline model performance with human performance on 1,000 random samples. ``Human UB'' stands for the upper bound on annotator performance on \datasetname. For details please refer to the main body.} \label{tab:human_perf}
\end{table}
\setlength{\tabcolsep}{6pt}

\section{Related Work}
Various recently-proposed large-scale QA datasets can be categorized in four categories.

\paragraph{Single-document datasets.} \squad{} \cite{rajpurkar2016squad, rajpurkar2018know} questions that are relatively simple because they usually require no more than one sentence in the paragraph to answer.

\paragraph{Multi-document datasets.} \triviaqa{} \cite{Joshi2017TriviaQA} and \searchqa{} \cite{dunn2017searchqa} contain question answer pairs that are accompanied with more than one document as the context.
This further challenges QA systems' ability to accommodate longer contexts.
However, since the supporting documents are collected after the question answer pairs with information retrieval, the questions are not guaranteed to involve interesting reasoning between multiple documents.

\paragraph{KB-based multi-hop datasets.}
Recent datasets like
\wikihop{} \cite{welbl2017constructing} and \complexwebq{} \cite{talmor2018web} explore different approaches of using pre-existing knowledge bases (KB) with pre-defined logic rules to generate valid QA pairs, to test QA models' capability of performing multi-hop reasoning.
The diversity of questions and answers is largely limited by the fixed KB schemas or logical forms.
Furthermore, some of the questions might be answerable by one text sentence due to the incompleteness of KBs.

\paragraph{Free-form answer-generation datasets.}
MS MARCO \cite{nguyen2016ms} contains 100k user queries from Bing Search with human generated answers. Systems generate free-form answers and are evaluated by automatic metrics such as ROUGE-L and BLEU-1. However, the reliability of these metrics is questionable because they have been shown to correlate poorly with human judgement \cite{novikova2017we}.


\section{Conclusions}

We present \datasetname{}, a large-scale question answering dataset aimed at facilitating the development of QA systems capable of performing explainable, multi-hop reasoning over diverse natural language. We also offer a new type of factoid comparison questions to test systems' ability to extract and compare various entity properties in text.

\section*{Acknowledgements}
This work is partly funded by the Facebook ParlAI Research Award. ZY, WWC, and RS are supported by a Google grant, the DARPA grant D17AP00001, the ONR grants N000141512791, N000141812861, and the Nvidia NVAIL Award.
SZ and YB are supported by Mila, Universit\'e de Montr\'eal.
PQ and CDM are supported by the National Science Foundation under Grant No.\ IIS-1514268. Any opinions, findings, and conclusions or recommendations expressed in this material are those of the authors and do not necessarily reflect the views of the National Science Foundation.

\bibliography{hotpotqa}

\begin{thebibliography}{19}
\expandafter\ifx\csname natexlab\endcsname\relax\def\natexlab#1{#1}\fi

\bibitem[{Chen et~al.(2017)Chen, Fisch, Weston, and Bordes}]{chen2017reading}
Danqi Chen, Adam Fisch, Jason Weston, and Antoine Bordes. 2017.
\newblock Reading {Wikipedia} to answer open-domain questions.
\newblock In \emph{Association for Computational Linguistics (ACL)}.

\bibitem[{Clark and Gardner(2017)}]{clark2017simple}
Christopher Clark and Matt Gardner. 2017.
\newblock Simple and effective multi-paragraph reading comprehension.
\newblock In \emph{Proceedings of the 55th Annual Meeting of the Association of
  Computational Linguistics}.

\bibitem[{Dunn et~al.(2017)Dunn, Sagun, Higgins, Guney, Cirik, and
  Cho}]{dunn2017searchqa}
Matthew Dunn, Levent Sagun, Mike Higgins, Ugur Guney, Volkan Cirik, and
  Kyunghyun Cho. 2017.
\newblock {SearchQA}: A new {Q\&A} dataset augmented with context from a search
  engine.
\newblock \emph{arXiv preprint arXiv:1704.05179}.

\bibitem[{Joshi et~al.(2017)Joshi, Choi, Weld, and
  Zettlemoyer}]{Joshi2017TriviaQA}
Mandar Joshi, Eunsol Choi, Daniel~S. Weld, and Luke Zettlemoyer. 2017.
\newblock {TriviaQA}: A large scale distantly supervised challenge dataset for
  reading comprehension.
\newblock In \emph{Proceedings of the 55th Annual Meeting of the Association
  for Computational Linguistics}.

\bibitem[{Liu et~al.(2018)Liu, Shen, Duh, and Gao}]{liu2017stochastic}
Xiaodong Liu, Yelong Shen, Kevin Duh, and Jianfeng Gao. 2018.
\newblock Stochastic answer networks for machine reading comprehension.
\newblock In \emph{Proceedings of the 56th Annual Meeting of the Association
  for Computational Linguistics}.

\bibitem[{Manning et~al.(2014)Manning, Surdeanu, Bauer, Finkel, Bethard, and
  McClosky}]{manning-EtAl:2014:P14-5}
Christopher~D. Manning, Mihai Surdeanu, John Bauer, Jenny Finkel, Steven~J.
  Bethard, and David McClosky. 2014.
\newblock The {Stanford} {CoreNLP} natural language processing toolkit.
\newblock In \emph{Association for Computational Linguistics (ACL) System
  Demonstrations}, pages 55--60.

\bibitem[{Miller et~al.(2017)Miller, Feng, Fisch, Lu, Batra, Bordes, Parikh,
  and Weston}]{miller2017parlai}
Alexander~H Miller, Will Feng, Adam Fisch, Jiasen Lu, Dhruv Batra, Antoine
  Bordes, Devi Parikh, and Jason Weston. 2017.
\newblock {ParlAI}: A dialog research software platform.
\newblock \emph{arXiv preprint arXiv:1705.06476}.

\bibitem[{Nguyen et~al.(2016)Nguyen, Rosenberg, Song, Gao, Tiwary, Majumder,
  and Deng}]{nguyen2016ms}
Tri Nguyen, Mir Rosenberg, Xia Song, Jianfeng Gao, Saurabh Tiwary, Rangan
  Majumder, and Li~Deng. 2016.
\newblock {MS MARCO}: A human generated machine reading comprehension dataset.
\newblock In \emph{Proceedings of the 30th Annual Conference on Neural
  Information Processing Systems (NIPS)}.

\bibitem[{Novikova et~al.(2017)Novikova, Du{\v{s}}ek, Curry, and
  Rieser}]{novikova2017we}
Jekaterina Novikova, Ond{\v{r}}ej Du{\v{s}}ek, Amanda~Cercas Curry, and Verena
  Rieser. 2017.
\newblock Why we need new evaluation metrics for {NLG}.
\newblock In \emph{Proceedings of the Conference on Empirical Methods in
  Natural Language Processing}.

\bibitem[{Pan et~al.(2017)Pan, Li, Zhao, Cao, Cai, and He}]{pan2017memen}
Boyuan Pan, Hao Li, Zhou Zhao, Bin Cao, Deng Cai, and Xiaofei He. 2017.
\newblock Memen: Multi-layer embedding with memory networks for machine
  comprehension.
\newblock \emph{arXiv preprint arXiv:1707.09098}.

\bibitem[{Rajpurkar et~al.(2018)Rajpurkar, Jia, and Liang}]{rajpurkar2018know}
Pranav Rajpurkar, Robin Jia, and Percy Liang. 2018.
\newblock Know what you don't know: Unanswerable questions for {SQuAD}.
\newblock In \emph{Proceedings of the 56th Annual Meeting of the Association
  for Computational Linguistics}.

\bibitem[{Rajpurkar et~al.(2016)Rajpurkar, Zhang, Lopyrev, and
  Liang}]{rajpurkar2016squad}
Pranav Rajpurkar, Jian Zhang, Konstantin Lopyrev, and Percy Liang. 2016.
\newblock {SQuAD}: 100,000+ questions for machine comprehension of text.
\newblock In \emph{Proceedings of the 2016 Conference on Empirical Methods in
  Natural Language Processing (EMNLP)}.

\bibitem[{Salant and Berant(2018)}]{salant2017contextualized}
Shimi Salant and Jonathan Berant. 2018.
\newblock Contextualized word representations for reading comprehension.
\newblock In \emph{Proceedings of the 16th Annual Conference of the North
  American Chapter of the Association for Computational Linguistics}.

\bibitem[{Seo et~al.(2017)Seo, Kembhavi, Farhadi, and
  Hajishirzi}]{seo2016bidirectional}
Minjoon Seo, Aniruddha Kembhavi, Ali Farhadi, and Hannaneh Hajishirzi. 2017.
\newblock Bidirectional attention flow for machine comprehension.
\newblock In \emph{Proceedings of the International Conference on Learning
  Representations}.

\bibitem[{Talmor and Berant(2018)}]{talmor2018web}
Alon Talmor and Jonathan Berant. 2018.
\newblock The web as a knowledge-base for answering complex questions.
\newblock In \emph{Proceedings of the 16th Annual Conference of the North
  American Chapter of the Association for Computational Linguistics}.

\bibitem[{Wang et~al.(2017)Wang, Yang, Wei, Chang, and Zhou}]{wang2017gated}
Wenhui Wang, Nan Yang, Furu Wei, Baobao Chang, and Ming Zhou. 2017.
\newblock Gated self-matching networks for reading comprehension and question
  answering.
\newblock In \emph{Proceedings of the 55th Annual Meeting of the Association
  for Computational Linguistics (Volume 1: Long Papers)}, volume~1, pages
  189--198.

\bibitem[{Welbl et~al.(2018)Welbl, Stenetorp, and
  Riedel}]{welbl2017constructing}
Johannes Welbl, Pontus Stenetorp, and Sebastian Riedel. 2018.
\newblock Constructing datasets for multi-hop reading comprehension across
  documents.
\newblock \emph{Transactions of the Association of Computational Linguistics}.

\bibitem[{Xiong et~al.(2018)Xiong, Zhong, and Socher}]{xiong2017dcn+}
Caiming Xiong, Victor Zhong, and Richard Socher. 2018.
\newblock {DCN+}: Mixed objective and deep residual coattention for question
  answering.
\newblock In \emph{Proceedings of the International Conference on Learning
  Representations}.

\bibitem[{Yang et~al.(2018)Yang, Zhang, Urbanek, Feng, Miller, Szlam, Kiela,
  and Weston}]{yang2017mastering}
Zhilin Yang, Saizheng Zhang, Jack Urbanek, Will Feng, Alexander~H Miller,
  Arthur Szlam, Douwe Kiela, and Jason Weston. 2018.
\newblock Mastering the dungeon: Grounded language learning by mechanical
  turker descent.
\newblock In \emph{Proceedings of the International Conference on Learning
  Representations}.

\end{thebibliography}
\bibliographystyle{acl_natbib_nourl}

\clearpage
\appendix

\section{Data Collection Details} \label{sec:collection_details}

\subsection{Data Preprocessing}

We downloaded the dump of English Wikipedia of October 1, 2017, and extracted text and hyperlinks with WikiExtractor.\footnote{\url{https://github.com/attardi/wikiextractor}}
 We use Stanford CoreNLP 3.8.0 \cite{manning-EtAl:2014:P14-5} for word and sentence tokenization.
 We use the resulting sentence boundaries for collection of supporting facts, and use token boundaries to check whether Turkers are providing answers that cover spans of entire tokens to avoid nonsensical partial-word answers.

\subsection{Further Data Collection Details}

\paragraph{Details on Curating Wikipedia Pages.}
To make sure the sampled candidate paragraph pairs are intuitive for crowd workers to ask high-quality multi-hop questions about, we manually curate 591 categories from the lists of popular pages by WikiProject.\footnote{\url{https://wiki.sh/y8qu}}
For each category, we sample $(a, b)$ pairs from the graph $G$ where $b$ is in the considered category, and manually check whether a multi-hop question can be asked given the pair $(a, b)$.
Those categories with a high probability of permitting multi-hop questions are selected.

\paragraph{Bonus Structures.} To incentivize crowd workers to produce higher-quality data more efficiently, we follow \citet{yang2017mastering}, and employ bonus structures. We mix two settings in our data collection process. In the first setting, we reward the top (in terms of numbers of examples) workers every 200 examples. In the second setting, the workers get bonuses based on their productivity (measured as the number of examples per hour).

\subsection{Crowd Worker Interface}

\begin{figure}
    \centering
    \includegraphics[width=.48\textwidth]{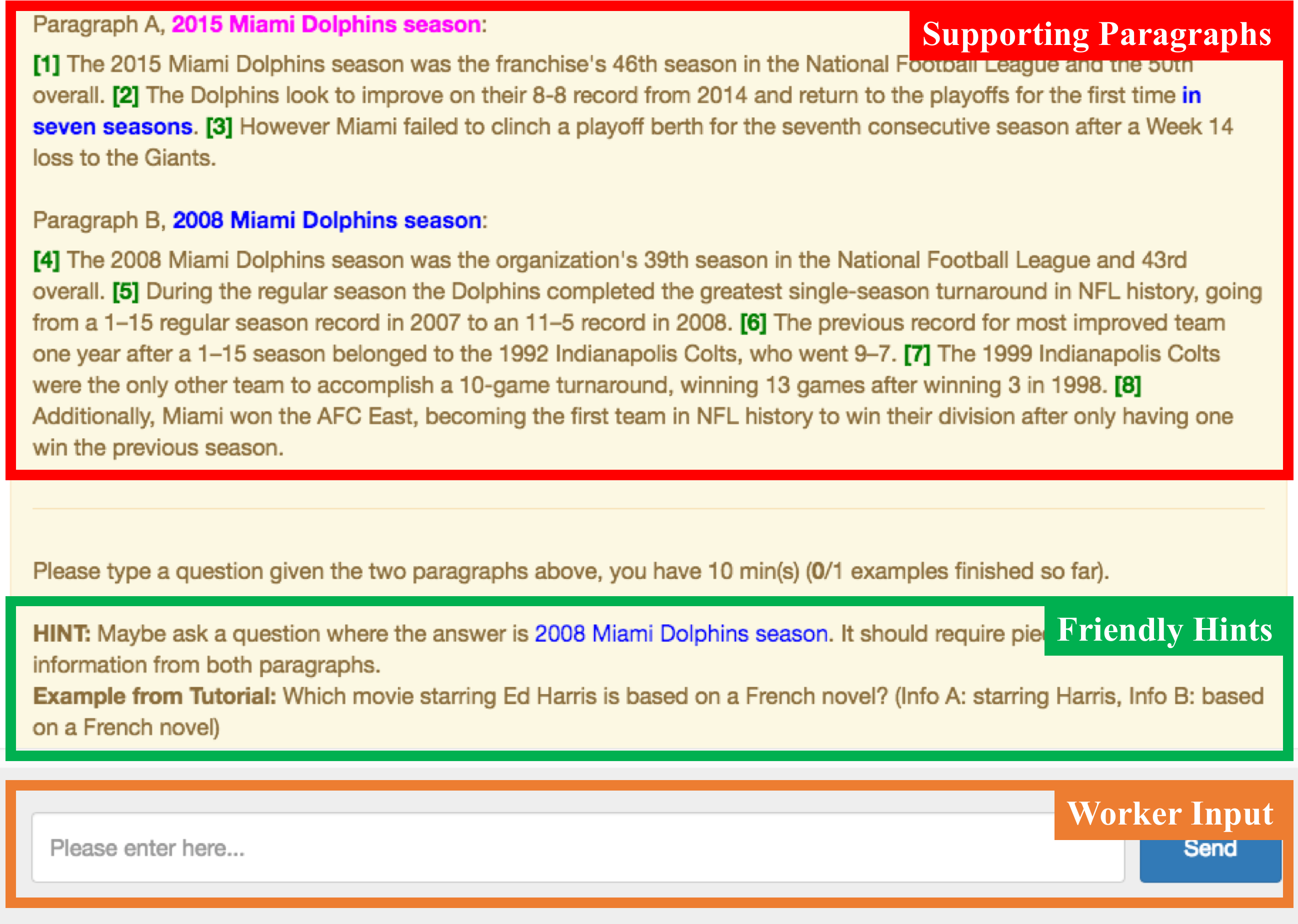}
    \caption{Screenshot of our worker interface on Amazon Mechanical Turk.} \label{fig:interface}
\end{figure}

Our crowd worker interface is based on \parlai{} \cite{miller2017parlai}, an open-source project that facilitates the development of dialog systems and data collection with a dialog interface.
We adapt \parlai{} for collecting question answer pairs by converting the collection workflow into a system-oriented dialog.
This allows us to have more control over the turkers input, as well as provide turkers with in-the-loop feedbacks or helpful hints to help Turkers finish the task, and therefore speed up the collection process.

Please see Figure \ref{fig:interface} for an example of the worker interface during data collection.

\section{Further Data Analysis} \label{sec:further_analysis}

To further look into the diversity of the data in \datasetname{}, we further visualized the distribution of question lengths in the dataset in Figure \ref{fig:question_lengths}.
Besides being diverse in terms of types as is show in the main text, questions also vary greatly in length, indicating different levels of complexity and details covered.

\begin{figure}
    \centering
    \pgfplotstableread[row sep=\\,col sep=&]{
	l & count \\
	10	&	11506\\
	15	&	40554\\
	20	&	34054\\
	25	&	18523\\
	30	&	8858\\
	35	&	4398\\
	40	&	2356\\
	45	&	1535\\
	50	&	1074\\
	55	&	768\\
	60	&	607\\
	65	&	485\\
	70	&	321\\
	75	&	224\\
	80	&	141\\
	85	&	94\\
	90	&	49\\
	95	&	44\\
	100	&	22\\
	105	&	13\\
	110	&	3\\
	115	&	6\\
	120	&	6\\
	125	&	3\\
	130	&	1\\
}\mydata

\pgfplotsset{compat=1.8,
	/pgfplots/ybar legend/.style={
		/pgfplots/legend image code/.code={%
			\draw[##1,/tikz/.cd,bar width=3.5pt,yshift=-0.2em,bar shift=0pt]
			plot coordinates {(0cm,0.8em)};},
	},
}

\begin{tikzpicture}[font=\small]
\begin{axis}[
ybar=.5pt,
xtick={10, 30, 50, 70, 90, 110, 130},
ymin=0, ymax=45000,
ytick={1e4, 2e4, 3e4, 4e4},
xlabel={Question Length (tokens)},
ylabel={Number of Examples},
height=6.5cm,
/pgf/bar width=3.5pt,
xtick align=inside,
ymajorgrids=true,
grid style=dashed,
]

\addplot[draw opacity=0, fill=blue!50] table[x=l,y=count]{\mydata};

\end{axis}
\end{tikzpicture}
    \caption{Distribution of lengths of questions in \datasetname{}.} \label{fig:question_lengths}
\end{figure}

\section{Full Wiki Setting Details} \label{sec:fullwiki_details}
\subsection{The Inverted Index Filtering Strategy}

\begin{algorithm}
    \small
        \caption{Inverted Index Filtering Strategy}
        \label{algo:fullwiki_filter}
    \begin{algorithmic}
        \STATE {\bfseries Input:} question text $q$, control threshold $N$, ngram-to-Wikidoc inverted index $\mathcal{D}$
        \STATE {\bfseries Inintialize:}
            \STATE Extract unigram + bigram set $r_q$ from $q$
            \STATE $N_{cand} = +\infty$
            \STATE $C_{gram} = 0$
        \WHILE {$N_{cands} > N$}
            \STATE $C_{gram} = C_{gram} + 1$
            \STATE Set $S_{overlap}$ to be an empty dictionary
            \FOR {$w \in r_q$}
                 \FOR {$d \in \mathcal{D}[w]$}
                      \IF {$d$ not in $S_{overlap}$}
                           \STATE $S_{overlap}[d] = 1$
                      \ELSE
                           \STATE $S_{overlap}[d] = S_{overlap}[d] + 1$
                      \ENDIF
                 \ENDFOR
            \ENDFOR
            \STATE $S_{cand} = \emptyset$
            \FOR {$d$ in $S_{overlap}$}
                \IF {$S_{overlap}[d] \geq C_{gram}$}
                    \STATE $S_{cand} = S_{cand} \cup \{d\}$
                \ENDIF
            \ENDFOR
            \STATE $N_{cands} = |S_{cand}|$
        \ENDWHILE
        \RETURN $S_{cand}$
    \end{algorithmic}
    \end{algorithm}
In the full wiki setting, we adopt an efficient
inverted-index-based filtering strategy for preliminary
candidate paragraph retrieval. We provide details in Algorithm \ref{algo:fullwiki_filter}, where
we set the control threshold $N = 5000$ in our experiments.
For some of the question $q$, its corresponding gold paragraphs may not
be included in the output candidate pool $S_{cand}$,
we set such missing gold paragraph's rank as $|S_{cand}| + 1$ during the evaluation, so
MAP and Mean Rank reported in this paper are upper bounds of their true values.

\subsection{Compare \textit{train-medium} Split to Hard Ones}
Table \ref{tab:fullwiki_ir_train} shows the comparison between \textit{train-medium} split
and hard examples like \textit{dev} and \textit{test} under retrieval metrics in full wiki setting.
As we can see, the performance gap between \textit{train-medium} split and its
\textit{dev}/\textit{test} is close, which implies that \textit{train-medium} split has a similar level of
difficulty as hard examples under the full wiki setting in which a retrieval model is necessary as the first processing step.
\begin{table}[H]
    \small
    \centering
    \begin{tabular}{lcccc}
        \toprule
        Set & MAP & Mean Rank & CorAns Rank\\
        \midrule
        train-medium & 41.89 & 288.19 & 82.76 \\
        dev & 42.79 & 304.30 & 97.93  \\
        test & 45.92 & 286.20 & 74.85 \\
        \bottomrule
    \end{tabular}
    \caption{Retrieval performance comparison on full wiki setting for \textit{train-medium}, \textit{dev} and \textit{test} with 1,000 random samples each. MAP and are in \%. Mean Rank averages over retrieval ranks of two gold paragraphs. CorAns Rank refers to the rank of the gold paragraph containing the answer.} \label{tab:fullwiki_ir_train}
\end{table}

\end{document}